%% file: paper.tex
\documentclass[sigconf]{acmart}

\usepackage[utf8]{inputenc} 
\usepackage[T1]{fontenc}    
\usepackage{hyperref}       
\usepackage{url}            
\usepackage{booktabs}       
\usepackage{amsfonts}       
\usepackage{nicefrac}       
\usepackage{microtype}      
\usepackage{xcolor}         
\usepackage{wrapfig}
\usepackage{graphicx}
\usepackage{subfigure}
\usepackage{balance}
\usepackage{amsmath}
\usepackage{amssymb}
\usepackage{mathtools}
\usepackage{amsthm}
\usepackage{algorithm}
\usepackage{algorithmic}
\usepackage{adjustbox}
\usepackage{makecell}
\usepackage{multirow}
\usepackage[capitalize,noabbrev]{cleveref}
\usepackage{balance}

\theoremstyle{plain}

\theoremstyle{definition}

\theoremstyle{remark}

\newcommand{\norm}[1]{\left\lVert#1\right\rVert}
\newcommand{\R}{\mathbb{R}}

\colorlet{Mycolor1}{green!10!orange}
\definecolor{Mycolor2}{HTML}{00F9DE}

\AtBeginDocument{%
  \providecommand\BibTeX{{%
    \normalfont B\kern-0.5em{\scshape i\kern-0.25em b}\kern-0.8em\TeX}}}

\copyrightyear{2025}
\acmYear{2025}
\setcopyright{cc}
\setcctype{by}
\acmConference[KDD '25] {Proceedings of the 31st ACM SIGKDD Conference on Knowledge Discovery and Data Mining V.2}{August 3--7, 2025}{Toronto, ON, Canada.}
\acmBooktitle{Proceedings of the 31st ACM SIGKDD Conference on Knowledge Discovery and Data Mining V.2 (KDD '25), August 3--7, 2025, Toronto, ON, Canada}
\acmDOI{10.1145/3711896.3737081}
\acmISBN{979-8-4007-1454-2/2025/08}
\begin{document}

\title{Physics-Guided Learning of Meteorological Dynamics\\ for Weather Downscaling and Forecasting}

\author{Yingtao Luo}
\authornote{Work done during summer internship at Alibaba Group.}
\affiliation{
  \institution{Carnegie Mellon University}
  \city{Pittsburgh}
  \country{USA}
  }
\email{yingtaol@andrew.cmu.edu}

\author{Shikai Fang}
\authornotemark[1]
\affiliation{
  \institution{Microsoft Research Asia}
  \city{Beijing}
  \country{China}
  }
\email{fangshikai@microsoft.com}

\author{Binqing Wu}
\authornotemark[1]
\affiliation{
  \institution{Zhejiang University}
  \city{Hangzhou}
  \country{China}
  }
\email{binqingwu@cs.zju.edu.cn}

\author{Qingsong Wen}
\authornote{Work done at Alibaba Group, and now affiliated with Squirrel Ai Learning, USA.}
\affiliation{
  \institution{Squirrel Ai Learning}
  \city{Bellevue}
  \country{USA}
  }
\email{qingsongedu@gmail.com}

\author{Liang Sun}
\authornote{To whom correspondence should be addressed.}
\affiliation{
  \institution{DAMO Academy, Alibaba Group}
  \city{Bellevue}
  \country{USA}
  }
\email{liang.sun@alibaba-inc.com}



\begin{abstract}
Weather forecasting is essential but remains computationally intensive and physically incomplete in traditional numerical weather prediction (NWP) methods. Deep learning (DL) models offer efficiency and accuracy but often ignore physical laws, limiting interpretability and generalization.
We propose \textit{PhyDL-NWP}, a physics-guided deep learning framework that integrates physical equations with latent force parameterization into data-driven models. It predicts weather variables from arbitrary spatiotemporal coordinates, computes physical terms via automatic differentiation, and uses a physics-informed loss to align predictions with governing dynamics.
\textit{PhyDL-NWP} enables resolution-free downscaling by modeling weather as a continuous function and fine-tunes pre-trained models with minimal overhead, achieving up to 170× faster inference with only 55K parameters. Experiments show that \textit{PhyDL-NWP} improves both forecasting performance and physical consistency.
\end{abstract}

\maketitle

\section{Introduction}
Weather prediction remains one of modern science's most complex and vital challenges, with the nonlinear interactions between various meteorological variables, the vast spatial and temporal scales involved, and the chaotic nature of weather systems. The first-principle approach to weather prediction, i.e., numerical weather prediction (NWP), relies on mathematical models of atmospheric and oceanic phenomena. NWP is often computationally intensive at high resolution, and many unresolved physical processes such as precipitation and radiation need to be represented by parameterization. Therefore, there has been a growing interest in using machine learning (ML) models for weather prediction. Deep learning models, trained by nearly 40 years of European Center for Medium-Range Weather
Forecasts (ECMWF) reanalysis v5 (ERA5) data~\citep{hersbach2020era5}, have demonstrated remarkable ability to capture complex nonlinear relationships for tasks such as weather forecasting~\citep{pathak2022fourcastnet, wu2023interpretable} and downscaling~\citep{vandal2017deepsd}. However, despite the recent success of ML techniques, the application of deep learning to weather prediction is not without challenges. Existing ML models do not incorporate established physical laws (e.g., fluid dynamics, thermodynamics) to ensure that the derived variables are consistent with these laws. 

Physics-Informed Neural Networks (PINN)~\citep{raissi2019physics} have gained prominence as alternatives to traditional numerical simulations, offering innovative approaches to weather prediction. However, weather prediction is inherently complex, involving numerous factors and processes that are influenced by local variations, boundary condition changes, small-scale phenomena like microclimates, and external forces. Many of these critical factors, which significantly affect first-principle equations, are often missing from existing datasets due to challenges in measurement and quantification. For instance, the PDE governing temperature evolution (see Table~\ref{tab:discovery}) includes thermal diffusivity, which is not available in typical weather datasets such as ERA5 and must instead be estimated through turbulence parameterizations. Other unavailable terms include vertical velocity, friction, etc. This poses a challenge to constructing complete and accurate physical models, thereby limiting the fidelity of physics-informed learning approaches in real-world settings.

In light of this challenge, we propose a novel framework, \textit{PhyDL-NWP}. This proposed paradigm first trains a neural network to predict weather conditions based on spatio-temporal coordinates, then leverages automatic differentiation for calculating partial differential equation (PDE) terms. Based on this, we construct a library of physics terms derived from first-principle equations that can be represented using available dataset variables. To address unresolved processes and missing variables, we adopt a parametrization strategy, introducing a latent force model as a parametrization term to capture the effects of physical forces not explicitly represented in the constructed PDEs.

This design enables \textit{PhyDL-NWP} to generate super-resolution weather data at arbitrary granularity, while simultaneously providing interpretable physical insights. Based on the constructed governing equations that represent the physical principles, we constrain and guide the optimization of deep learning models, enhancing model performance across datasets with varying climates and sources. In extensive experiments across 17 baselines and four datasets, we demonstrate that the learned parameterized PDEs align closely with the desired physical equations. The physics-guided models consistently outperform their vanilla counterparts.

In short, our contributions are summarized as follows:
\begin{itemize}
\item We propose a novel physics-guided learning framework \textit{PhyDL-NWP} that completes weather equations using latent force parameterization to inform deep learning models of the underlying physical mechanism of meteorology. 
\item \textit{PhyDL-NWP} directly provides a novel way to model weather data in an online learning manner with unlimited granularity. Weather downscaling can be done by simply feeding any continuous coordinates to the model, without the need for coarse-granular data as input during inference. 
\item \textit{PhyDL-NWP} provides an effective paradigm for weather forecasting by forcing the prediction to align with the parameterized weather equations via a physics-guided loss.
\item \textit{PhyDL-NWP} is extremely efficient and can be directly used to fine-tune any pre-trained weather forecasting model. It can be up to 170 times faster in inference than a standalone model, with only up to 55 thousand parameters. 
\item The state-of-the-art performance of \textit{PhyDL-NWP} is evaluated using both reanalysis and real-world observational datasets, spanning global and local scales. \textit{PhyDL-NWP} shows consistency with the underlying physics in a variety of weather variables.
\end{itemize}

\section{Related Work}
\label{related}

In the area of weather prediction, Numerical Weather Prediction (NWP)~\citep{lorenc1986analysis, bauer2015quiet} is the current mainstream method. It uses mathematical models of the atmosphere and oceans, such as partial differential equations (PDE), to predict future weather based on current weather conditions. Some notable NWP models include European Centre for Medium-Range Weather Forecasts (ECMWF)\footnote{https://www.ecmwf.int/}, Global Forecast System (GFS)\footnote{https://www.ncei.noaa.gov/products/weather-climate-models/global-forecast}, etc. NWP can forecast weather in the medium range but usually involves extensive computation. For example, ECMWF operates one of the largest supercomputer complexes in Europe. 

Recently, deep learning has emerged as another promising solution to weather forecasting~\citep{hu2021improved, price2023gencast} and downscaling~\citep{vandal2017deepsd, park2022downscaling} tasks. These deep learning models~\citep{wang2019deep, han2022short} rely on different neural architectures such as LSTM~\citep{li2022numerical}, CNN~\citep{weyn2020improving}, GNN~\citep{lin2022conditional} and Transformer~\citep{wu2023interpretable} to capture the evolving dynamics and correlation across space and time. Many large models have emerged in recent years. For example, ClimaX~\citep{nguyen2023climax}, GraphCastNet~\citep{lam2022graphcast}, ClimateLearn~\citep{nguyen2023climatelearn}, FengWu~\citep{chen2023fengwu}, Pangu-Weather~\citep{bi2023accurate} all use backbones such as the Vision Transformer (ViT), UNet and autoencoders, for training a large model for weather forecasting. WeatherBench~\citep{rasp2020weatherbench} benchmarks the use of pre-training techniques for weather forecasting. In addition, FourcastNet~\citep{pathak2022fourcastnet} leverages the adaptive Fourier neural operator (AFNO)~\citep{li2020fourier, guibas2021adaptive} to treat weather as a latent PDE system. 
In recent years, a few studies such as NeuralGCM~\citep{kochkov2024neural} and WeatherGFT~\citep{xu2024generalizing} have started to explore the integration of underlying physical mechanisms~\citep{vermaclimode} in weather prediction. However, there are still no explicit efforts to use parameterization to ensure the completeness of the primitive equations based on the data.

In addition, spatio-temporal modeling based on deep learning~\citep{liang2025foundation} has thrived in recent years. Many previous works also approach the weather prediction task from the perspective of spatio-temporal modeling~\citep{han2021joint}. 
Dynamical systems modeling involves the formulation of systems whose states evolve over time. Given the governing equations, physics-informed approaches~\citep{raissi2019physics, karniadakis2021physics} use the physics mechanism to enhance the dynamical systems. In the absence of governing equations, the identification of physical equations~\citep{luo2022learning, chen2022symbolic, luo2023physics} is proposed to provide insights with respect to the laws of physics. 

\section{Methodology}
\subsection{Problem Definition}

We study a spatiotemporal weather dataset, denoted as 
\begin{align} \label{eq:def1}
\text{u} = [u_1(x, y, t), \ldots, u_h(x, y, t)]. 
\end{align}
Using the physics expression, this dataset comprises $h$ distinct weather variable fields (such as temperature and pressure), each related to specific input coordinates $(x, y, t)$. Here, $x \in [1, \ldots, n]$ and $y \in [1, \ldots, m]$ represent spatial coordinates, while $t \in [1, \ldots, T]$ corresponds to the temporal dimension.

This dataset can be alternatively expressed as a sequence of spatial “images” $\mathbb{X} = [X_1, \ldots, X_T]$, with each “image” $X_i$ being a tensor in $\mathbb{R}^{n \times m \times h}$, encapsulating the spatial and weather-factor dimensions at each time point.

We focus on two primary tasks based on this dataset:
\label{problem}
\begin{itemize}
    \item \textbf{Weather Downscaling}: The goal is to generate super-resolution weather “images” $\mathbb{Y} = [Y_1, \ldots, Y_T]$ from $\mathbb{X}$, where each $Y_i$ is a tensor in $\mathbb{R}^{n' \times m' \times h}$. The challenge is to derive this detailed data from the original, coarser dataset $\mathbb{X}$, with the dimensions $n' > n$ and $m' > m$. 
    \item \textbf{Weather Forecasting}: This task aims to find a model $g$ to predict future weather conditions for a duration of $r$ hours, represented as $[\mathbb{X}]_{i+1}^{i+r} = [X_{i+1}, \ldots, X_{i+r}]$. These predictions are based on observed data from the preceding $s+1$ hours, $[\mathbb{X}]_{i-s}^{i} = [X_{i-s}, \ldots, X_i]$, for each time instance $i$. That is, $g([\mathbb{X}]_{i-s}^{i})=[\mathbb{X}]_{i+1}^{i+r}$.
\end{itemize}

\begin{figure*}
\centering
\includegraphics[width=1\linewidth]{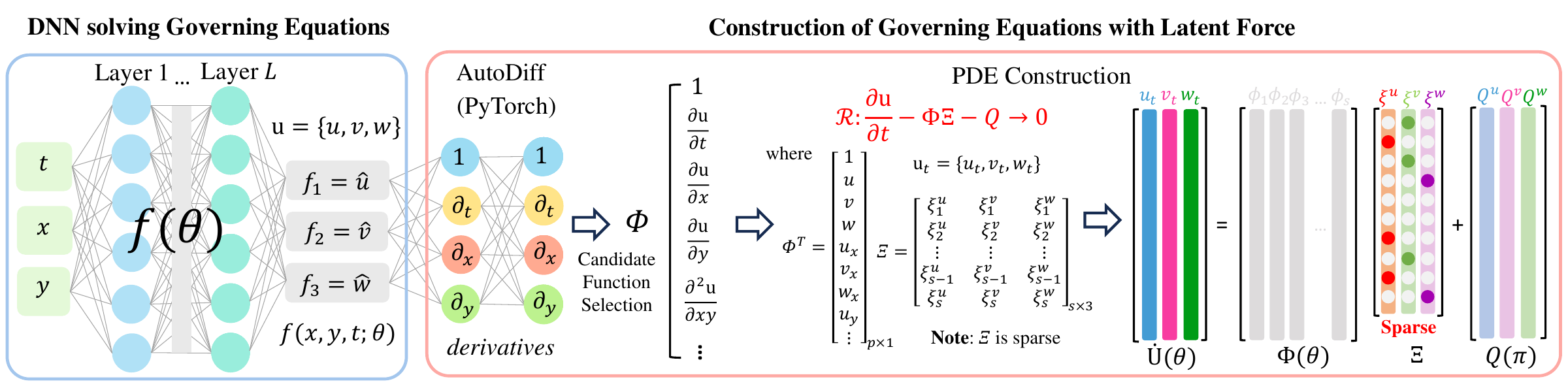}\vspace{-4mm}
\caption{Schematic diagram of \textit{PhyDL-NWP} for downscaling. First, given a continuous input coordinate $(x,y,t)$, the surrogate model $f_\theta$ approximates the weather data. Then, based on PyTorch's auto-differentiation and the existing meteorology theory, we calculate the derivatives for the construction of physical mechanisms driven by PDE. Last, based on linear regression, we learn the PDE that fits the weather data well to provide physical guidance.}\vspace{-2mm}
\label{fig:1}
\end{figure*}

\subsection{Meteorology Dynamics Representation}

In this section, we aim to learn the physical mechanism represented by PDE with parameterization to fit the weather data. A typical partial differential equation (PDE) with parametrization terms has the following form:
\begin{align} \label{eq:def2}
\frac{\partial \text{u}}{\partial t}
=Q_\pi(x,y,t)+\Phi(\text{u}) \Xi
=Q_\pi(x,y,t) + \sum_{i=1}^{p}\phi(\text{u})_i\xi_i,
\end{align}
where $p$ denotes the number of PDE terms in the equation, each $\phi(\text{u})_i \in [1, \text{u},\frac{\partial \text{u}}{\partial x},\frac{\partial \text{u}}{\partial y},\frac{\partial^2 \text{u}}{\partial x^2},..., \text{u}\frac{\partial \text{u}}{\partial x}, ...]$ denotes a PDE term in the equation (see examples in Sec. \ref{sec:discovery}), with the set of $\xi_i$ as the coefficients. $Q_\pi(x,y,t)$ denotes the latent force modeled by a neural network that cannot be represented by $\Phi(\text{u})$ explicitly, as a supplement to missing variables unavailable in the weather dataset, such as friction. Examples of Eq. \ref{eq:def2} can be seen in Table \ref{tab:discovery}, where a physical equation is composed of both explicit PDE terms and latent force parameterization. 

\subsection{Continuous Weather Downscaling}
We develop a surrogate weather variable model $\hat{\text{u}}=f_{\theta}(x,y,t)$, which takes the spatio-temporal coordinates as input and predicts the $h$ weather variables. The schematic diagram of the proposed surrogate model is shown in Fig. \ref{fig:1}. Both $f_{\theta}$ and $Q_\pi(x,y,t)$ are designed as feedforward neural networks that are commonly used in the PINN literature~\citep{raissi2019physics, karniadakis2021physics}. 
The joint optimization of $f_{\theta}(x,y,t)$, $\phi(\hat{\text{u}})i$, and $Q\pi(x,y,t)$ allows the model to simultaneously learn to predict weather variables and approximate the underlying physical dynamics. 
Once $f_{\theta}$ achieves sufficient accuracy, its gradients via automatic differentiation enable the accurate calculation of PDE terms. These inferred terms, together with $Q_\pi$, in turn guide the training of $f_\theta$, ensuring physical consistency in the learned mapping.

This framework supports continuous weather downscaling in an online learning manner: once trained, $f_\theta$ can be queried with any new coordinate $(x, y, t)$ for real-time prediction. Unlike traditional downscaling methods that rely on discrete pairs of low- and high-resolution training data, \textit{PhyDL-NWP} treats weather data as a continuous function over space-time. This enables arbitrary-resolution inference without requiring coarse-resolution inputs or pre-defined grid structures. Downscaling is performed simply by evaluating $Y = f_{\theta}(x', y', t)$ at any desired fine-grained coordinates in real time, as described in Fig. \ref{fig:1}, where $x' \in \R^{n'}, y' \in \R^{m'}$ can be any interpolation within the observed spatial domain, enabling super-resolution modeling with minimal inference cost. 

The overall loss function $\mathcal{L}_{\text{D}}$ for weather downscaling is:
\begin{align} \label{eq:def3}
\mathcal{L}_{\text{D}}(\theta, \Xi, \pi) &=  \mathcal{L}_{\text{data}}(\theta) + \alpha \mathcal{L}_{\text{phy}}(\theta, \Xi, \pi),
\end{align}
where
\begin{align} \label{eq:def4}
\mathcal{L}_{\text{data}} &= \frac{1}{\text{nmT}} \sum_{x, y, t} \norm{f_\theta-\text{u}}_2^2, \\
 \label{eq:def5}
\mathcal{L}_{\text{phy}} &= \frac{1}{\text{n}'\text{m}'\text{T}'} \sum_{x', y', t'}  \norm{\frac{\partial f_\theta}{\partial t}-\Phi(f_\theta) \Xi-Q_\pi}_2^2.
\end{align}
Here, data loss measures how well $f_\theta$ approximates $u$ well on the weather data, and physical loss measures how well the learned equation fits the weather data. The two regularization losses prevent the overfitting of explicit PDE terms $f_\theta$ and the latent force $Q_\pi$. 

\begin{figure*}
\centering
\includegraphics[width=1\linewidth]{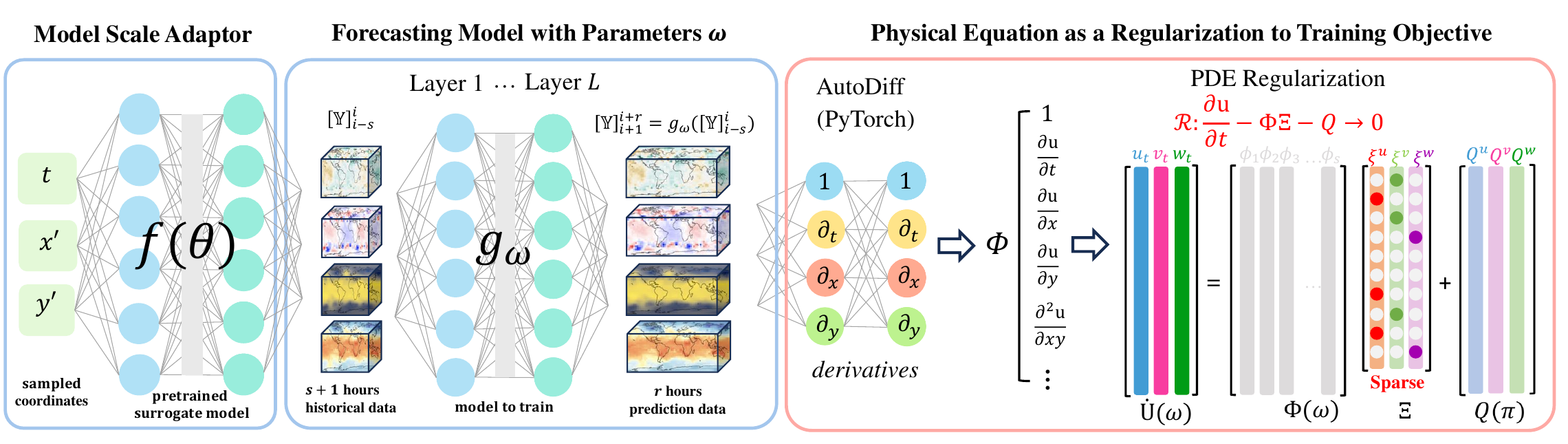}\vspace{-4mm}
\caption{Schematic diagram of \textit{PhyDL-NWP} for forecasting. We first use pre-trained surrogate model for weather downscaling to perform data augmentation, which is a necessity for aligning weather data resolution in the forecasting model. Then, we take the augmented historical data and use a pre-trained state-of-the-art forecasting model to predict future data. Based on the spatio-temporal coordinates of the predicted data, we add a physics loss to recover the previously learned PDE. }\vspace{-2mm}
\label{fig:2}
\end{figure*}

\subsection{Physics-Guided Fine-Tuning for Pre-Trained Weather Forecasting Models}
In theory, $f_\theta (x,y,t)$ can also take in future data coordinates and produce the extrapolation directly. However, as $f_\theta (x,y,t)$ is only trained on historical weather data, while anywhere outside the bounds of where the model was trained is completely unknown. Empirically, we find that $f_\theta$ alone does not exhibit strong extrapolation or forecasting performance.
To improve the extrapolation ability, instead of using $f_\theta$ directly, we propose to take advantage of the learned physical mechanism represented by $\Xi$ and $Q_\pi$ to improve another forecasting model $g_\omega$, which takes historical spatiotemporal data and predicts the future. 

Moreover, once $f_\theta$ is trained, it can generate weather variables at arbitrary spatiotemporal coordinates $(x', y', t)$, enabling flexible control over the resolution of historical data used to train or fine-tune $g_\omega$. This allows seamless integration with any pre-trained forecasting model by aligning spatiotemporal granularity as needed.

The overall framework of weather forecasting is depicted in Fig. \ref{fig:2}. To calculate differential terms efficiently, we propose to use finite difference (FD) approximations on super-resolution data from weather downscaling, which is extremely fast, instead of training a surrogate model for every time frame $i$. 
The overall loss function $\mathcal{L}_{\text{F}}$ for weather forecasting is:
\begin{align} \label{eq:def6}
\mathcal{L}_{\text{F}}(\omega) &=  \mathcal{L}_{\text{data}}(\omega) + \beta \mathcal{L}_{\text{phy}}(\omega),
\end{align}
where
\begin{align} \label{eq:def7}
\mathcal{L}_{\text{data}}(\omega) &= \frac{1}{\text{nmr} \cdot q} \sum_{i=s+1}^{T-r} \norm{g_\omega([\mathbb{Y}]_{i-s}^{i})-[\mathbb{Y}]_{i+1}^{i+r}}_2^2, \\
\label{eq:def8}
\mathcal{L}_{\text{phy}}(\omega) &= \frac{1}{\text{nmr} \cdot q} \sum_{i=s+1}^{T-r} \norm{\frac{\partial g_\omega([\mathbb{Y}]_{i-s}^{i})}{\partial t}-p}_2^2,
\end{align}
with $q = \text{T-r-s-1}$$, $\quad $p = \Phi(g_\omega([\mathbb{Y}]_{i-s}^{i})) \Xi + Q_\pi$. 
Note that $\theta$, $\Xi$, and $\pi$ are already learned during the downscaling beforehand. $\alpha, \beta, \sigma_1, \sigma_2$ are all hyperparameters to balance the different loss terms. As long as the downscaling model $f_\theta$ is accurate, the recovered physics offers a globally consistent constraint that enhances the generalization of $g_\omega$ without adding substantial model complexity. The size of the parameters as well as the inference speed of the different models are summarized in Table~\ref{tab:param}. The number of parameters of \textit{PhyDL-NWP} is much smaller than other models, thus it is much faster to perform forward/backward propagation on $f_\theta$ and $g_\omega$. 

\begin{table*}[h]
\centering
  \caption{Comparison of the number of model parameters and running speed for weather forecasting on WeatherBench dataset. \textit{PhyDL-NWP} is a light-weighted and efficient plug-and-play module, while others are standalone models. \textit{PhyDL-NWP} is about 55$\sim$170 times faster and 10$\sim$3600 times lighter than a standalone model.}\vspace{-2mm}
  \label{tab:param}
  \begin{adjustbox}{width=0.98\textwidth}
  \begin{tabular}{c| c| c| c| c| c| c| c| c| c |c}
    \toprule
    \hline
    Model/Module &  \textbf{PhyDL-NWP} & BiLSTM & Hybrid-CBA & ConvLSTM & AFNO & MTGNN & MegaCRN & ClimaX & FourcastNet & GraphCast \\
    \hline
    Number of parameters &  \textbf{55K} & 171M & 198M & 678K & 520K & 1.6M & 580K & 107M & 73M & 36M \\
    \hline
    Time cost per epoch  &  \textbf{7.8s} & 11.9min & 15.6min & 7.1min & 9.4min & 9.9min & 8.5min & 22.2min & 16.5min & 13.6min \\
    \hline
    \bottomrule
  \end{tabular}
  \end{adjustbox}\vspace{-2mm}
\end{table*}

\section{Experiments}
We conduct both the forecasting and downscaling performance comparisons. All experiments were carried out on four NVIDIA A100 graphical cards. Only the performances on the test sets at the optimal performance on the validation sets are reported. The maximum training epochs are 50. Every result is the average of three independent trainings under different random seeds. We select a few representative weather variables and the average of all in the tables for visualization. The average of all variables reflects the overall performance. We use two commonly used metrics~\citep{bi2023accurate} for evaluation: Root Mean Square Error (RMSE) and Anomaly Correlation Coefficient (ACC). For both downscaling and forecasting, we split by 8:1:1 for train/validation/test datasets in chronological order. Code to implement PhyDL-NWP is available in GitHub\footnote{\url{https://github.com/yingtaoluo/PhyDL-NWP}}.

\begin{table*}[t]
\centering
  \caption{The RMSE comparison of weather downscaling for Huadong dataset. Bold fonts mark the best performances, and underlines mark the second-best performances. The Improv shows the percentage of improvement over the base model, which is statistically significant as measured by t-test with p-value$<0.01$.}\vspace{-2mm}
  \label{tab:huadong}
  \begin{adjustbox}{width=0.72\textwidth}
\begin{tabular}{c| c c| c c| c c| c c| c c}
    \toprule
    \hline
    \multirow{2}*{Model} & \multicolumn{2}{c|}{100m Wind (U)} & \multicolumn{2}{c|}{10m Wind (U)} & \multicolumn{2}{c|}{Temperature} & \multicolumn{2}{c|}{Surface Pressure} & \multicolumn{2}{c}{Average} \\
    ~ & 2x & 4x & 2x & 4x & 2x & 4x & 2x & 4x & 2x & 4x \\
    \hline
    Bicubic & 1.687 & 1.765 & 1.215 & 1.272 & 1.714 & 1.848 & 0.818 & 1.220 & 1.515 & 1.654  \\
    EDSR & 1.145 & 1.176 & 1.020 & 1.113 & 1.217 & 1.275 & 0.460 & 0.552 & 1.068 & 1.156  \\
    ResDeepD & \underline{1.092} & \underline{1.111} & 1.003 & 1.079 & 1.182 & \underline{1.204} & \underline{0.301} & \underline{0.317} & \underline{1.010} & \underline{1.043} \\
    RCAN & 1.169 & 1.199 & \underline{0.808} & \underline{1.038} & 1.219 & 1.259 & 0.572 & 0.609 & 1.092 & 1.144 \\
    FSRCNN & 1.197 & 1.202 & 1.090 & 1.126 & 1.198 & 1.233 & 0.430 & 0.560 & 1.093 & 1.149  \\
    YNet & 1.116 & 1.125 & 0.947 & 1.103 & 1.192 & 1.226 & 0.467 & 0.575 & 1.062 & 1.125 \\
    DeepSD & 1.205 & 1.216 & 1.020 & 1.117 & 1.218 & 1.265 & 0.454 & 0.591 & 1.087 & 1.149\\
    GINE & 1.126 & 1.285 & 0.875 & 1.069 & \underline{1.166} & 1.235 & 0.350 & 0.363 & 1.036 & 1.101  \\
    PhyDL-NWP & \textbf{0.973} & \textbf{0.970} & \textbf{0.696} & \textbf{0.693} & \textbf{0.905} & \textbf{0.904} & \textbf{0.211} & \textbf{0.216} & \textbf{0.794} & \textbf{0.789} \\
    \hline
    Improv & 10.9\% & 12.7\% & 13.9\% & 33.2\% & 22.4\% & 24.9\% & 29.9\% & 31.9\% & 20.1\% & 24.6\% \\
    \hline
    \bottomrule
  \end{tabular}
  \end{adjustbox}
\end{table*}

\subsection{Downscaling Performance Comparison}
We evaluate the effectiveness of \textit{PhyDL-NWP} and other baseline models for weather downscaling on a real-world dataset Huadong, which is derived from the European Centre for Medium-Range Weather Forecasts (ECMWF) operational forecast (HRES) and reanalysis (ERA5) archive. It comprises a grid of $64 \times 44$ cells, with each cell having a grid size of 0.25 degrees in both latitude and longitude. More data details can be found in Appendix \ref{app:data_description}. Since most previous studies on weather downscaling can only handle the downscaling of the two spatial dimensions, for the sake of comparison, we also only report the performance of \textit{PhyDL-NWP} on spatial downscaling in Table \ref{tab:huadong}. We perform 2x and 4x downscaling tasks with 0.5 and 1 degrees resolutions, respectively. To facilitate this, the 0.25-degree HRES data undergoes linear interpolation to generate the requisite 0.5-degree and 1-degree input data. We compare our model against the Bicubic interpolation, FSRCNN~\citep{passarella2022reconstructing}, ResDeepD~\citep{sharma2022resdeepd}, EDSR~\citep{jiang2022efficient}, RCAN~\citep{yu2021deep}, YNet~\citep{liu2020climate}, DeepSD~\citep{vandal2017deepsd}, and GINE~\citep{park2022downscaling}. For the deep learning baselines, channel-wise normalization is performed for efficiency. Unlike prior methods that rely on fine-granular labels for supervision, \textit{PhyDL-NWP} learns from coarse-granular inputs alone by aligning with physical principles, enabling it to perform super-resolution without labeled training outputs. Details about baselines can be found in Appendix \ref{app-baseline}.

From Table \ref{tab:huadong}, we can conclude that \textit{PhyDL-NWP} provides a significant improvement up to 20.1\% to 24.6\% on average over RMSE against the baseline models. Well-recognized deep learning models like FSRCNN and YNet achieve much worse results, which could be because most models only consider the downscaling of spatial dimensions, neglecting the patterns in the temporal dimension. Moreover, weather data is multivariable, and the spatio-temporal dependencies are complex, making it difficult to recover the ground-truth information without global modeling. Furthermore, we find that the RMSE for 2x and 4x resolutions is close. Since \textit{PhyDL-NWP} can provide infinite resolution results given continuous coordinates, we believe that it will be accurate for higher resolution downscaling, based on this evidence. Moreover, \textit{PhyDL-NWP} can easily perform downscaling in the temporal dimension. We only experiment on spatial dimension to align with existing models.

\begin{table*}[t]
\centering
  \caption{Model comparison of seven-day weather forecasting for real-measurement Ningbo dataset. The Improv shows the percentage of improvement over the base model, which is statistically significant as measured by t-test with p-value$<0.01$.}\vspace{-2mm}
  \label{tab:7dayningbo}
  \begin{adjustbox}{width=0.75\textwidth}
  \begin{tabular}{c| c c| c c| c c| c c| c c}
    \toprule
    \hline
    \multirow{2}*{Model} & \multicolumn{2}{c|}{100m Wind} & \multicolumn{2}{c|}{10m Wind} & \multicolumn{2}{c|}{Humidity} & \multicolumn{2}{c|}{Temperature} & \multicolumn{2}{c}{Average} \\
    ~ & RMSE$\downarrow$ & ACC$\uparrow$ & RMSE$\downarrow$ & ACC$\uparrow$ & RMSE$\downarrow$ & ACC$\uparrow$ & RMSE$\downarrow$ & ACC$\uparrow$ & RMSE$\downarrow$ & ACC$\uparrow$ \\
    \hline
    NWP & 0.892 & 0.606 & 0.875 & 0.581 & 0.932 & \textbf{0.699} & 0.422 & 0.910 & 0.868 & \textbf{0.587}  \\
    PINN & \textbf{0.622} & 0.520 & \textbf{0.605} & 0.489 & 0.835 & 0.443 & 0.657 & 0.727 & 0.652 & 0.427 \\
    PINO & 0.640 & 0.504 & 0.602 & 0.516 & 0.609 & 0.538 & 0.477 & 0.838 & 0.626 & 0.452 \\
    \hline
    Bi-LSTM-T & 0.666 & 0.588 & 0.704 & 0.562 & 0.576 & 0.597 & 0.472 & 0.876 & 0.601 & 0.443  \\
    Bi-LSTM-T+ & 0.635 & \textbf{0.649} & \underline{0.664} & \underline{0.621} & 0.550 & 0.672 & 0.442 & 0.903 & 0.571 & 0.485 \\
    Improv & 4.65\% & 10.4\% & 5.68\% & 10.5\% & 4.51\% & 12.6\% & 6.36\% & 3.08\% & 5.00\% & 9.48\% \\
    \hline
    Hybrid-CBA & 0.674 & 0.568 & 0.717 & 0.550 & 0.590 & 0.595 & 0.460 & 0.865 & 0.617 & 0.431  \\
    Hybrid-CBA+ & 0.641 & 0.637 & 0.680 & 0.609 & 0.572 & 0.657 & 0.411 & 0.906 & 0.586 & 0.474 \\
    Improv & 4.90\% & 12.1\% & 5.16\% & 10.7\% & 3.05\% & 10.4\% & 10.7\% & 4.74\% & 5.02\% & 9.98\% \\
    \hline
    ConvLSTM & 0.701 & 0.524 & 0.732 & 0.535 & 0.572 & 0.602 & 0.489 & 0.858 & 0.636 & 0.418 \\
    ConvLSTM+ & 0.658 & 0.587 & 0.699 & 0.607 & 0.550 & 0.671 & 0.454 & 0.891 & 0.596 & 0.463 \\
    Improv & 6.13\% & 12.0\% & 4.51\% & 13.5\% & 3.85\% & 11.5\% & 7.16\% & 3.85\% & 5.97\% & 10.8\% \\
    \hline
    AFNO & 0.659 & 0.592 & 0.710 & 0.546 & 0.528 & 0.584 & 0.429 & 0.894 & 0.599 & 0.465 \\
    AFNO+ & \underline{0.625} & \underline{0.648} & 0.669 & \textbf{0.630} & \underline{0.500} & \underline{0.695} & \underline{0.397} & \textbf{0.929} & \underline{0.556} & \underline{0.530} \\
    Improv & 5.16\% & 9.46\% & 5.78\% & 15.4\% & 5.30\% & 19.0\% & 7.46\% & 3.91\% & 7.18\% & 14.0\% \\
    \hline
    MTGNN & 0.685 & 0.566 & 0.720 & 0.538 & 0.521 & 0.589 & 0.434 & 0.887 & 0.597 & 0.457 \\
    MTGNN+ & 0.657 & 0.629 & 0.672 & 0.613 & \textbf{0.489} & 0.679 & \textbf{0.388} & \underline{0.918} & \textbf{0.555} & 0.514 \\
    Improv & 4.09\% & 11.1\% & 6.67\% & 13.9\% & 6.14\% & 15.3\% & 10.6\% & 3.49\% & 7.04\% & 12.5\% \\
    \hline
    MegaCRN & 0.698 & 0.520 & 0.734 & 0.535 & 0.544 & 0.595 & 0.492 & 0.866 & 0.621 & 0.426  \\
    MegaCRN+ & 0.667 & 0.591 & 0.684 & 0.600 & 0.521 & 0.666 & 0.458 & 0.907 & 0.590 & 0.477 \\
    Improv & 4.44\% & 13.7\% & 6.81\% & 12.1\% & 4.23\% & 11.9\% & 6.91\% & 4.73\% & 5.00\% & 12.0\% \\
    \hline
    \bottomrule
  \end{tabular}
  \end{adjustbox}
  \vspace{-3mm}
\end{table*}

\begin{figure*}[t]
\centering
\includegraphics[width=0.85\linewidth]{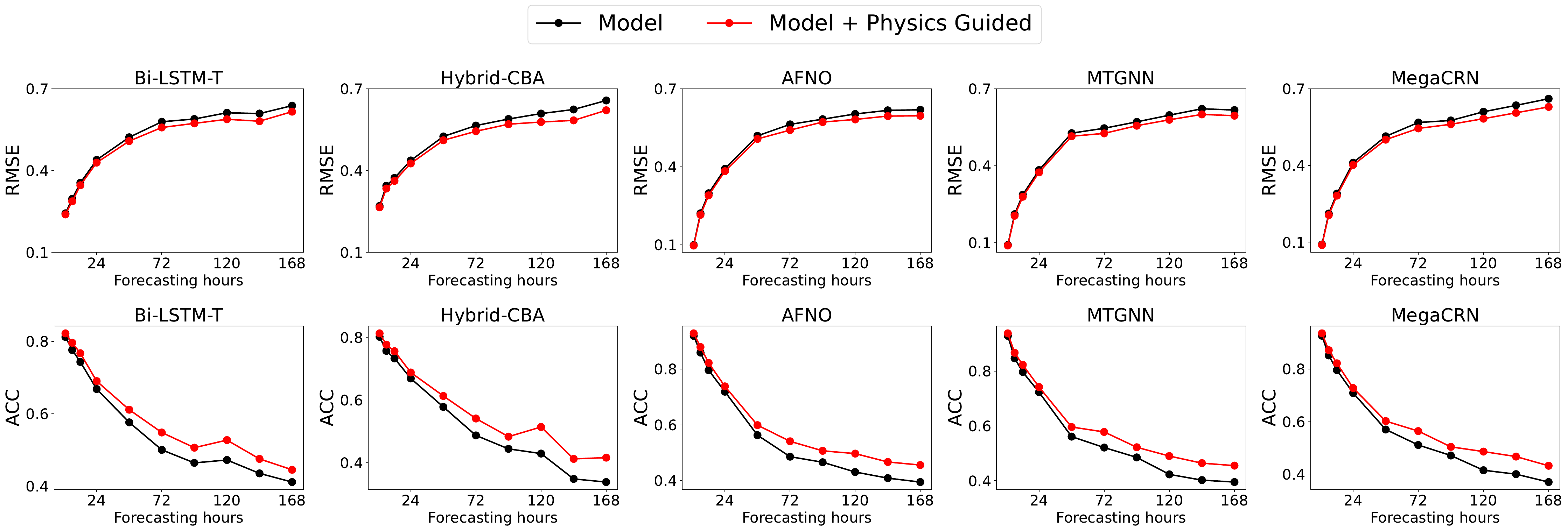}\vspace{-5mm}
\caption{Model comparison in Ningxia dataset before and after physics guidance for a variety of forecasting ranges on the average of all weather variables. }
\label{fig:3}
\end{figure*}

\subsection{Forecasting Performance Comparison}
We evaluate the effectiveness of \textit{PhyDL-NWP} for weather forecasting on two real-world reanalysis datasets derived from ERA5\footnote{\url{https://www.ecmwf.int/en/forecasts/datasets}}: Ningxia and WeatherBench\footnote{\url{https://mediatum.ub.tum.de/1524895}}. In addition to reanalysis data, we also test on a more accurate real-world measurement (observational) dataset Ningbo, where meteorological factors are directly collected from the Ningbo Meteorological Bureau\footnote{\url{http://zj.cma.gov.cn/dsqx/nbsqxj/}}. Ningbo and Ningxia cover two different terrains and climate types in 0.25 degrees resolution, while WeatherBench covers the global weather in 5.625 degrees resolution. On each grid in both datasets, we select the most important observational weather information for evaluation. 
See more details in Appendix \ref{app:data_description}. We perform multiple experiments based on the length of future prediction, ranging from one hour to seven days. Due to the GPU memory limitation, we use the eight weather variables of only ten hours in the past to predict all eight weather variables in the future. 

There are five kinds of baseline models in comparison, including: (1) Physics-based models: NWP, PINN~\citep{raissi2019physics}, PINO~\citep{li2024physics}; (2) Meteorological models: Bi-LSTM-T~\citep{yang2022correcting}, Hybrid-CBA~\citep{han2022short}; (3) Vision models: ConvLSTM~\citep{shi2015convolutional}, FourcastNet~\citep{pathak2022fourcastnet} based on AFNO~\citep{guibas2021adaptive}, ClimaX~\citep{nguyen2023climax}; (4) Spatio-temporal graph models: MTGNN~\citep{wu2020connecting}, MegaCRN~\citep{jiang2023spatio}, GraphCast~\citep{lam2022graphcast}. Some baseline models are slightly modified to adapt to the multi-step prediction setting and/or specific modeling resolution, with details described in the Appendix \ref{app-baseline}. Besides deep learning models, we also compare these models with the Numerical Weather Prediction (NWP) results provided by ECMWF IFS and the Physical-Informed Neural Network (PINN)~\citep{raissi2019physics} based on the PDEs learned by \textit{PhyDL-NWP}. We denote the baseline models as \textbf{BaseModels} and incorporate them with \textit{PhyDL-NWP} as \textbf{BaseModels+}. 

\begin{table*}[t]
\centering
\caption{Model comparison of global weather forecasting up to two days for the WeatherBench dataset. All the reported results are averaged after three runs. BaseModel+ denotes the original model with \textit{PhyDL-NWP} module.}\vspace{-2mm}
\label{tab:weatherbench}
\begin{adjustbox}{width=0.8\textwidth}
\begin{tabular}{c|c||cc|cc||cc|cc||cc|cc}
  \toprule
  \hline
  \multirow{2}{*}{Variable} & \multirow{2}{*}{Hours} & \multicolumn{2}{c|}{ClimaX} & \multicolumn{2}{c||}{ClimaX+} & \multicolumn{2}{c|}{FourcastNet} & \multicolumn{2}{c||}{FourcastNet+} & \multicolumn{2}{c|}{GraphCast} & \multicolumn{2}{c}{GraphCast+} \\
   & & RMSE$\downarrow$ & ACC$\uparrow$ & RMSE$\downarrow$ & ACC$\uparrow$ & RMSE$\downarrow$ & ACC$\uparrow$ & RMSE$\downarrow$ & ACC$\uparrow$ &
   RMSE$\downarrow$ & ACC$\uparrow$ &
   RMSE$\downarrow$ & ACC$\uparrow$ \\
  \hline
  \multirow{5}{*}{t2m} & 6  & 1.46 & 0.92 & 1.13 & 0.98 
                       & 1.27 & 0.99 & 1.01 & 0.99  
                       & 0.40 & 0.99 & 0.39 & 0.99  \\
                       & 12 & 1.58 & 0.91 & 1.25 & 0.96
                       & 1.48 & 0.98 & 1.12 & 0.99  
                       & 0.47 & 0.99 & 0.46 & 0.99  \\
                       & 18 & 1.75 & 0.90 & 1.42 & 0.95 
                       & 1.63 & 0.98 & 1.27 & 0.99  
                       & 0.52 & 0.99 & 0.50 & 0.99  \\
                       & 24 & 1.90 & 0.88 & 1.58 & 0.94 
                       & 1.69 & 0.96 & 1.40 & 0.98 
                       & 0.59 & 0.99 & 0.56 & 0.99  \\
                       & 48 & 2.80 & 0.84 & 2.34 & 0.92
                       & 2.26 & 0.94 & 1.90 & 0.97 
                       & 0.74 & 0.98 & 0.72 & 0.99  \\
                       
  \hline
  \multirow{5}{*}{t}   & 6  & 1.32 & 0.95 & 1.02 & 0.98 
                       & 1.15 & 0.99 & 0.99 & 0.99  
                       & 0.39 & 0.99 & 0.39 & 0.99  \\
                       & 12 & 1.66 & 0.94 & 1.28 & 0.98
                       & 1.36 & 0.99 & 1.17 & 0.99  
                       & 0.46 & 0.99 & 0.45 & 0.99  \\
                       & 18 & 1.87 & 0.92 & 1.48 & 0.97
                       & 1.53 & 0.99 & 1.35 & 0.99  
                       & 0.53 & 0.99 & 0.51 & 0.99  \\
                       & 24 & 2.16 & 0.91 & 1.66 & 0.96 
                       & 1.66 & 0.98 & 1.52 & 0.99  
                       & 0.59 & 0.99 & 0.57 & 0.99  \\
                       & 48 & 2.94 & 0.86 & 2.11 & 0.95
                       & 1.94 & 0.97 & 1.70 & 0.99  
                       & 0.80 & 0.99 & 0.77 & 0.99  \\
  \hline
  \multirow{5}{*}{z}   & 6  & 207.6 & 0.93 & 128.5 & 0.97 
                       & 142.3 & 0.96 & 100.8 & 0.99  
                       & 44.1 & 0.99 & 44.0 & 0.99  \\
                       & 12 & 222.3 & 0.90 & 159.9 & 0.96 
                       & 217.2 & 0.89 & 126.6 & 0.99  
                       & 47.6 & 0.99 & 47.2 & 0.99  \\
                       & 18 & 268.7 & 0.87 & 197.6 & 0.95 
                       & 255.0 & 0.74 & 166.2 & 0.98  
                       & 50.6 & 0.99 & 49.5 & 0.99  \\
                       & 24 & 305.5 & 0.84 & 224.1 & 0.94 
                       & 304.2 & 0.71 & 203.2 & 0.97  
                       & 78.4 & 0.98 & 75.7 & 0.99  \\
                       & 48 & 497.2 & 0.77 & 292.4 & 0.92 
                       & 477.6 & 0.62 & 278.0 & 0.95  
                       & 118.6 & 0.98 & 112.5 & 0.98  \\
  \hline
  \multirow{5}{*}{u10} & 6  & 1.56 & 0.90 & 1.28 & 0.94 
                       & 1.39 & 0.93 & 1.12 & 0.95  
                       & 0.50 & 0.98 & 0.50 & 0.98 \\
                       & 12 & 1.98 & 0.89 & 1.73 & 0.94 
                       & 1.88 & 0.92 & 1.69 & 0.94  
                       & 0.53 & 0.98 & 0.53 & 0.98 \\
                       & 18 & 2.20 & 0.89 & 1.94 & 0.93 
                       & 2.10 & 0.90 & 1.88 & 0.93  
                       & 0.57 & 0.98 & 0.56 & 0.98 \\
                       & 24 & 2.46 & 0.85 & 2.15 & 0.92 
                       & 2.36 & 0.89 & 2.09 & 0.92  
                       & 0.75 & 0.97 & 0.73 & 0.98 \\
                       & 48 & 2.91 & 0.78 & 2.46 & 0.88 
                       & 2.79 & 0.88 & 2.36 & 0.90  
                       & 1.24 & 0.96 & 1.16 & 0.97 \\
  \hline
  \multirow{5}{*}{v10} & 6  & 1.78 & 0.88 & 1.37 & 0.94 
                       & 1.55 & 0.94 & 1.22 & 0.94 
                       & 0.52 & 0.98 & 0.52 & 0.98 \\
                       & 12 & 1.99 & 0.86 & 1.52 & 0.93 
                       & 1.81 & 0.90 & 1.39 & 0.93 
                       & 0.55 & 0.98 & 0.55 & 0.98 \\
                       & 18 & 2.35 & 0.85 & 1.74 & 0.92 
                       & 2.11 & 0.88 & 1.63 & 0.92 
                       & 0.58 & 0.98 & 0.57 & 0.98 \\
                       & 24 & 2.66 & 0.83 & 2.08 & 0.90 
                       & 2.40 & 0.85 & 1.96 & 0.91 
                       & 0.79 & 0.97 & 0.76 & 0.98 \\
                       & 48 & 3.74 & 0.70 & 2.49 & 0.87 
                       & 3.06 & 0.80 & 2.25 & 0.89 
                       & 1.36 & 0.96 & 1.24 & 0.97 \\
  \hline
  \bottomrule
\end{tabular} 
\end{adjustbox}
\end{table*}

\begin{table*}
\centering
  \caption{Model comparison of seven-day medium-range weather forecasting for Ningxia dataset.}
  \label{tab:7dayningxia}
  \begin{adjustbox}{width=0.72\textwidth}
\begin{tabular}{c| c c| c c| c c| c c| c c}
    \toprule
    \hline
    \multirow{2}*{Model} & \multicolumn{2}{c|}{100m Wind(U)} & \multicolumn{2}{c|}{10m Wind(U)} & \multicolumn{2}{c|}{Temperature} & \multicolumn{2}{c|}{Surface pressure} & \multicolumn{2}{c}{Average} \\
    ~ & RMSE$\downarrow$ & ACC$\uparrow$ & RMSE$\downarrow$ & ACC$\uparrow$ & RMSE$\downarrow$ & ACC$\uparrow$ & RMSE$\downarrow$ & ACC$\uparrow$ & RMSE$\downarrow$ & ACC$\uparrow$ \\
    \hline
    NWP & 0.968 & 0.521 & 0.933 & 0.514 & \textbf{0.319} & \textbf{0.844} & 0.325 & 0.961 & 0.901 & \textbf{0.526}  \\
    PINN & \textbf{0.697} & 0.470 & \textbf{0.681} & 0.437 & 0.635 & 0.654 & 0.494 & 0.904 & 0.666 & 0.387  \\
    \hline
    Bi-LSTM-T & 0.822 & 0.502 & 0.804 & 0.485 & 0.583 & 0.525 & 0.160 & 0.961 & 0.638 & 0.411  \\
    Bi-LSTM-T+ & \underline{0.798} & \textbf{0.545} & \underline{0.777} & \underline{0.520} & 0.560 & 0.584 & 0.156 & \underline{0.965} & 0.616 & 0.445 \\
    Improv & 2.92\% & 8.57\% & 3.36\% & 7.22\% & 4.28\% & 11.2\% & 2.50\% & 0.42\% & 3.45\% & 8.27\% \\
    \hline
    Hybrid-CBA & 0.842 & 0.456 & 0.819 & 0.445 & 0.652 & 0.430 & \underline{0.150} & 0.964 & 0.657 & 0.338  \\
    Hybrid-CBA+ & 0.801 & \underline{0.536} & 0.790 & 0.509 & 0.563 & 0.589 & \textbf{0.149} & \textbf{0.966} & 0.621 & 0.416 \\
    Improv & 4.87\% & 17.5\% & 3.54\% & 14.4\% & 13.7\% & 37.0\% & 0.67\% & 0.21\% & 5.48\% & 18.8\% \\
    \hline
    ConvLSTM & 0.865 & 0.429 & 0.848 & 0.408 & 0.592 & 0.499 & 0.175 & 0.959 & 0.656 & 0.364 \\
    ConvLSTM+ & 0.826 & 0.477 & 0.814 & 0.472 & 0.520 & 0.619 & 0.170 & 0.955 & 0.622 & 0.419 \\
    Improv & 4.51\% & 11.2\% & 4.01\% & 15.7\% & 12.2\% & 24.0\% & 2.86\% & -0.42\% & 5.18\% & 15.1\% \\
    \hline
    AFNO & 0.856 & 0.436 & 0.838 & 0.421 & 0.501 & 0.571 & 0.153 & 0.962 & 0.619 & 0.395 \\
    AFNO+ & 0.823 & 0.505 & 0.808 & 0.498 & \underline{0.466} & 0.693 & 0.153 & 0.956 & \underline{0.596} & \underline{0.456} \\
    Improv & 3.86\% & 15.8\% & 3.58\% & 18.3\% & 6.99\% & 17.9\% & 0.00\% & -0.31\% & 3.72\% & 15.4\% \\
    \hline
    MTGNN & 0.835 & 0.484 & 0.820 & 0.465 & 0.502 & 0.526 & 0.162 & 0.958 & 0.617 & 0.395 \\
    MTGNN+ & 0.810 & 0.525 & 0.792 & \textbf{0.521} & 0.469 & \underline{0.677} & 0.160 & 0.959 & \textbf{0.595} & 0.455 \\
    Improv & 2.99\% & 8.47\% & 3.41\% & 12.0\% & 6.57\% & 28.7\% & 1.96\% & 0.10\% & 3.57\% & 15.2\% \\
    \hline
    MegaCRN & 0.840 & 0.455 & 0.824 & 0.432 & 0.646 & 0.487 & 0.188 & 0.958 & 0.661 & 0.370  \\
    MegaCRN+ & 0.809 & 0.510 & 0.793 & 0.485 & 0.598 & 0.600 & 0.183 & 0.954 & 0.629 & 0.432 \\
    Improv & 4.64\% & 12.1\% & 3.76\% & 12.3\% & 7.43\% & 23.2\% & 2.66\% & -0.42\% & 4.84\% & 16.8\% \\
    \hline
    \bottomrule
  \end{tabular}
  \end{adjustbox}
\end{table*}

\subsubsection{Regional Weather Forecasting}
In particular, the result of regional forecasting details with seven days are reported in Table \ref{tab:7dayningbo} in the main manuscript and Table \ref{tab:7dayningxia} in the appendix, as it represents the model's capability of long-term medium-range regional weather prediction. The vanilla PINN does not seem to be effective, while the improvement provided by \textit{PhyDL-NWP} is consistently significant. For Ningbo dataset, the overall improvement on the average of all weather variables is up to \textbf{7.18\% over RMSE} and \textbf{14.0\% over ACC}; for Ningxia dataset, the overall improvement on that is up to \textbf{5.48\% over RMSE} and \textbf{18.8\% over ACC}. All the results are \textbf{statistically significant}. Furthermore, we find that NWP is good in ACC, while being the worst in RMSE. Deep learning models, on the other hand, greatly outperform NWP in RMSE, showing great advantage in modeling capacity.

To understand the holistic properties of \textit{PhyDL-NWP}, we conduct detailed analyses on the Ningxia dataset. First, the comparison of different models for different forecasting ranges is visualized in Fig. \ref{fig:3}. \textbf{BaseModels+} excels \textbf{BaseModels} and NWP at all time steps. As the forecasting range increases, the deep learning performance decreases quickly. The improvement provided by \textit{PhyDL-NWP}, however, is increasing in the forecasting range, which highlights its unique advantages of guiding models for long-term forecasting.

\begin{figure*}[t]
    \centering
    \includegraphics[width=0.9\linewidth]{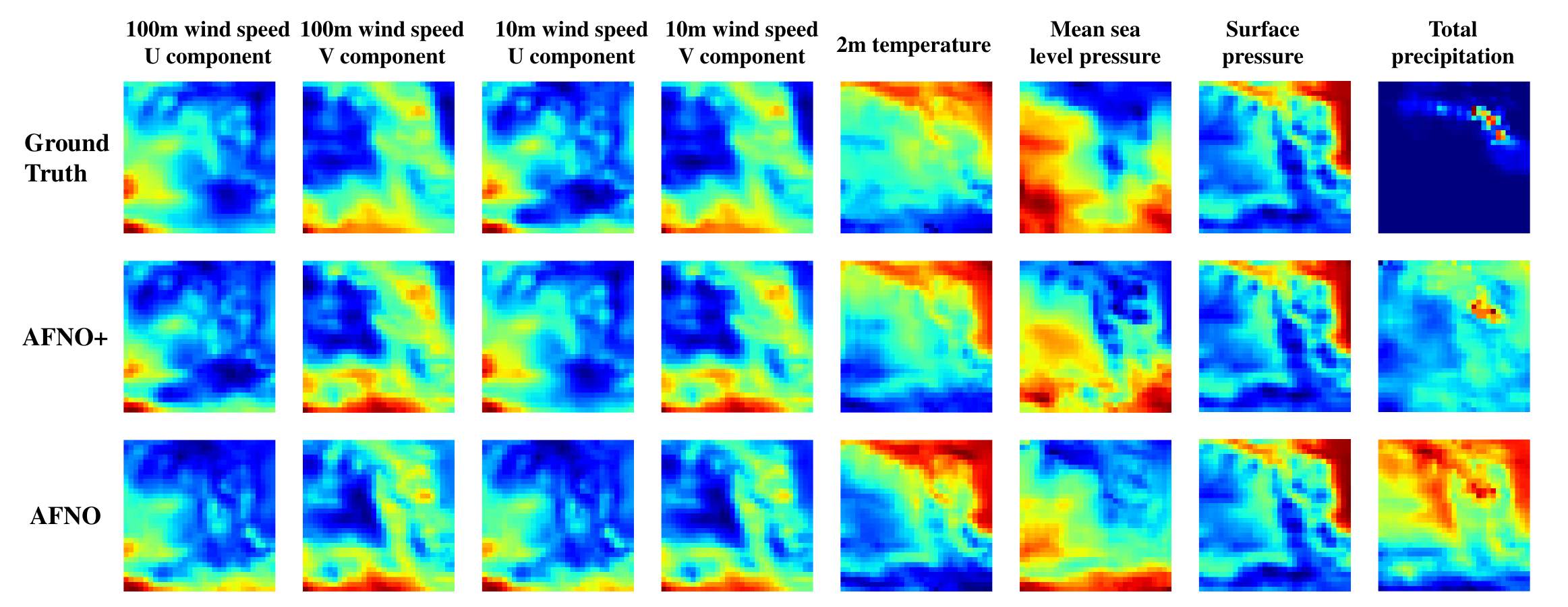}\vspace{-2mm}
    \caption{Example of comparison of 7-day weather forecast results in Ningxia dataset. AFNO+ are closer to the ground truth. }
    \label{fig:4}
    \vspace{-1mm}
\end{figure*}

In addition, based on AFNO from FourcastNet, we visualize its forecasting comparison at the 10-th time frame in Fig.~\ref{fig:4}. \textit{PhyDL-NWP} clearly improves the performance of the existing deep model, making the forecasting results closer to the ground truth. Specifically, “Total precipitation” is not a typical physical quantity and is hard to predict due to lack of effective physical equation. While the vanilla AFNO is a total disaster, AFNO+ provides information about which local areas receive concentrated rainfall. This effectively demonstrated the success of our implemented parameterization strategy with the latent force model.  

\subsubsection{Global Weather Forecasting}
Besides regional weather forecasting, we also test \textit{PhyDL-NWP} on the global weather forecasting task, which is the benchmark for a lot of recent works. The results with WeatherBench dataset for global weather forecasting are reported in Table \ref{tab:weatherbench}. In comparison, \textit{PhyDL-NWP} shows a lot more improvements over baseline models in the coarser-granular WeatherBench dataset, with \textbf{statistically significant improvement over RMSE}. It is also worth noting that, even for the state-of-the-art GraphCast model, there is still an improvement. Moreover, as reported in Table \ref{fig:1}, \textit{PhyDL-NWP} module is extremely light-weighted and efficient, with a time cost that is \textbf{55$\sim$170 times faster} than the base forecasting models. The integration of physics and deep learning is clearly demonstrated in this study.

\subsection{Meteorology Dynamics Interpretation}
\label{sec:discovery}

\begin{table}
\centering
  \caption{Comparison of learned dynamics against ground truth in literature. \textcolor{Mycolor1}{Orange color} marks terms unavailable from the data, which are represented by latent force $Q$.}\vspace{-2mm}
  \label{tab:discovery}
  \begin{adjustbox}{width=1\linewidth}
  \begin{tabular}{c| c }
    \toprule
    \hline
    Temperature & Learned vs. True PDE  \\
    \midrule
    Learned  & $\frac{\partial T}{\partial t} = - U_{10}\frac{\partial T}{\partial x} - V_{10}\frac{\partial T}{\partial y} + \textcolor{Mycolor1}{Q}$  \\
    Ground Truth & $\frac{\partial T}{\partial t} = - U\frac{\partial T}{\partial x} - V\frac{\partial T}{\partial y} - \textcolor{Mycolor1}{W \frac{\partial T}{\partial z}} + \textcolor{Mycolor1}{k \frac{\partial^2 T}{\partial z^2}} + \textcolor{Mycolor1}{H}$    \\
    \midrule
    Wind Velocity & Learned vs. True PDE  \\
    \midrule
    Learned  & $\frac{\partial U_{10}}{\partial t} = - U_{10}\frac{\partial U_{10}}{\partial x} - V_{10}\frac{\partial U_{10}}{\partial y} + \textcolor{Mycolor1}{Q}$  \\
    Ground Truth & $\frac{\partial U}{\partial t} = - U\frac{\partial U}{\partial x} - V\frac{\partial U}{\partial y} - \textcolor{Mycolor1}{W\frac{\partial U}{\partial z}} - \textcolor{Mycolor1}{\frac{1}{\rho} \frac{\partial p}{\partial x}} + \textcolor{Mycolor1}{\nu \Delta \textbf{u}} + \textcolor{Mycolor1}{F_{f_x}}$ \\
    \midrule
    Surface Pressure & Learned vs. True PDE  \\
    \midrule
    Learned  & $\frac{\partial^2 p}{\partial t^2} = \frac{\partial^2 p}{\partial x^2} + \frac{\partial^2 p}{\partial y^2} + \textcolor{Mycolor1}{Q}$  \\
    Ground Truth & $\frac{\partial^2 p}{\partial t^2} = \frac{\partial^2 p}{\partial x^2} + \frac{\partial^2 p}{\partial y^2} + \textcolor{Mycolor1}{\frac{\partial^2 p}{\partial z^2}}$  \\
        \midrule
    Humidity & Learned vs. True PDE  \\
    \midrule
    Learned  & $\frac{\partial q}{\partial t} = -U\frac{\partial q}{\partial x} - V\frac{\partial q}{\partial y} +  \frac{\partial^2 q}{\partial x^2} +  \frac{\partial^2 q}{\partial y^2} + \textcolor{Mycolor1}{Q}$  \\
    Ground Truth & $\frac{\partial q}{\partial t} = - U\frac{\partial q}{\partial x} - V\frac{\partial q}{\partial y} - \textcolor{Mycolor1}{W\frac{\partial q}{\partial z}} + \textcolor{Mycolor1}{k \nabla^2 q} + \textcolor{Mycolor1}{S_q}$  \\
    \hline
    \bottomrule
  \end{tabular}
  \end{adjustbox}
\end{table}

To understand how \textit{PhyDL-NWP} is grounded on laws of physics, we compare the learned dynamics by \textit{PhyDL-NWP} from Huadong dataset with the basic equations of NWP~\citep{stull1988introduction}. These equations originate from conservation of mass, energy, and momentum. \textit{PhyDL-NWP} will explicitly use the terms that appear in the first-principle equations and can be represented based on the dataset, while modeling the rest of the dynamics using latent force parameterization. As shown in Table \ref{tab:discovery}, our latent parameterization strategy is proposed considering that many terms cannot be modeled explicitly, due to missing records in the data or difficulty to measure. 

\begin{figure}
\centering
\includegraphics[width=0.54\linewidth]{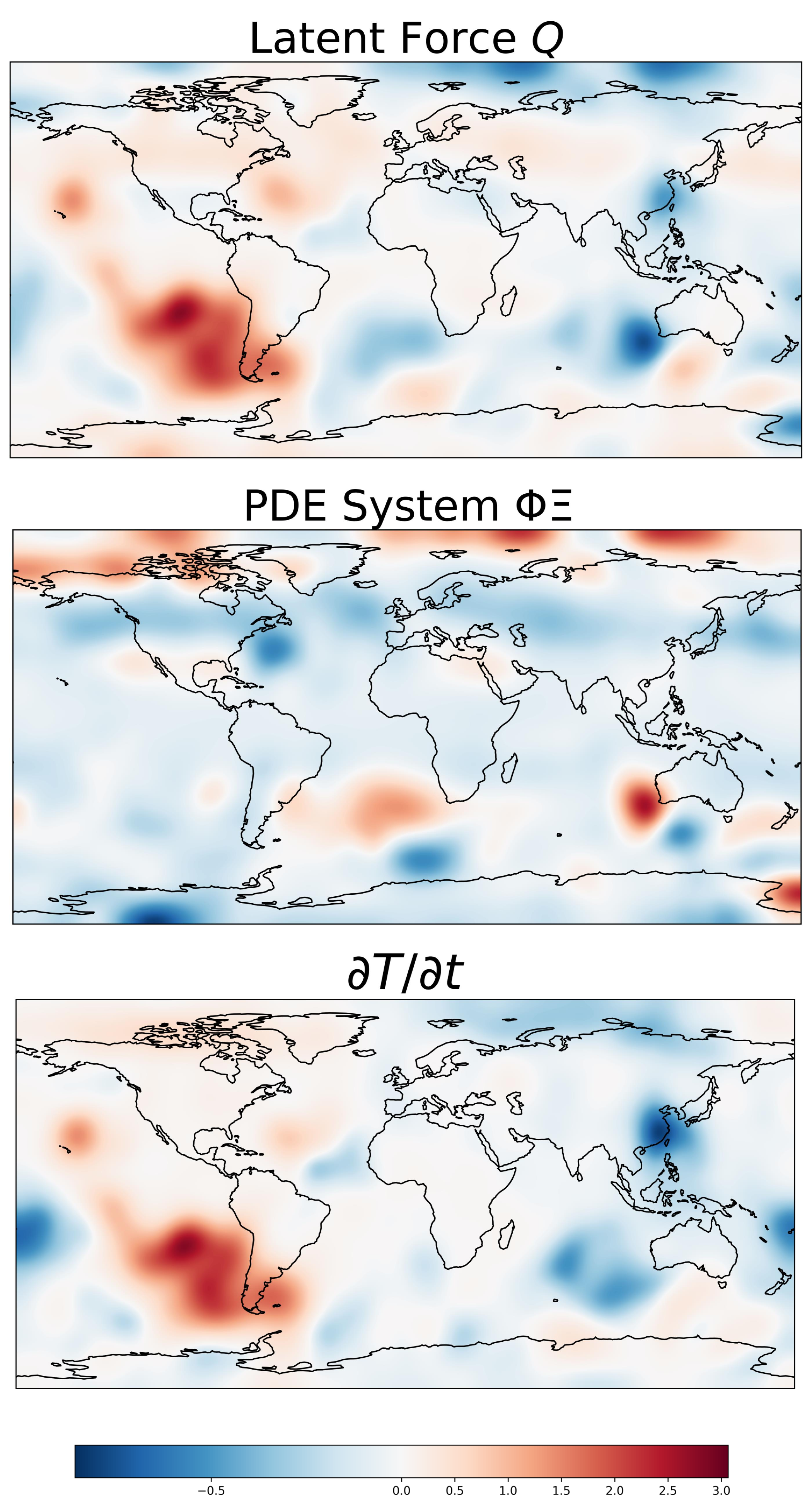}\vspace{-2mm}
\caption{The latent force and PDE for the temperature variation in WeatherBench dataset. }
\label{fig:weatherbench}
\vspace{-1mm}
\end{figure}

For the modeling of temperature, the basic equation originates from the 3D convection-diffusion equation for heat transfer. Temperature $T$ is influenced by advection (movement of heat due to fluid flow, represented by the velocity components $U$, $V$ and a vertical component $W$ along height $z$), diffusion (spread of heat due to thermal diffusivity $k$), and a heat source $H$. In addition, $t$ denotes time, $x$ and $y$ denote space, $U_{10}$ represents the wind components at the heights of 10m. 
An example of meteorology dynamics of temperature (t) of WeatherBench dataset in the year of 2018 is shown in Fig. \ref{fig:weatherbench}. We found that $Q$ and $\Phi \Xi$ substitute each other well and the combination generally matches $\frac{\partial T}{\partial t}$.

For wind velocity, the compared basic equation originates from the 3D Navier-Stokes equations. The velocity is influenced by convective acceleration, the gradient of pressure $p$ with fluid density $\rho$, the diffusion of momentum due to viscosity $\nu$ for velocity of all directions $\textbf{u}$, and external force $F_{f_x}$ along the $x$ direction (such as friction). It is obvious that the WeatherBench dataset does not provide all the necessary variables to complete the equation. Our latent force model will address this problem well. 
Similarly, for surface pressure, the learned equation aligns with the 3D wave equation while the missing vertical component is for the latent force $Q$ to capture. For humidity, the equation originates from the conservation of mass, and combines the advection, diffusion and source terms, where $q$ is the humidity, and $S_q$ is the source term for humidity (e.g., evaporation, condensation). We approximate precipitation using a parameterized PDE for humidity as a tractable surrogate that captures its dominant drivers.

\section{Conclusion}
We introduce \textit{PhyDL-NWP}, a light-weighted physics-guided deep learning framework for weather downscaling and forecasting. By incorporating parameterized physical dynamics through latent force modeling, \textit{PhyDL-NWP} enables continuous, resolution-free predictions while ensuring consistency with governing equations. It can augment and fine-tune existing forecasting models with minimal overhead, significantly improving both accuracy and physical plausibility. Extensive experiments across diverse datasets demonstrate its effectiveness, scalability, and interpretability, positioning \textit{PhyDL-NWP} as a practical and generalizable module for modern meteorological modeling.

\clearpage

\bibliographystyle{ACM-Reference-Format}
\balance
\bibliography{reference}

\newpage
\clearpage
\input{appendix}

\end{document}

%% file: appendix.tex
\newpage
\appendix
\section{Appendix}
\subsection{Dataset Description}
\label{app:data_description}

\begin{algorithm}
    \caption{Physics-Guided Learning of Numerical Weather Prediction (PhyDL-NWP).}
    \label{alg}
    \begin{algorithmic}
        \STATE \textbf{Require} Weather data $\text{u} = [u_1(x,y,t), \ldots,u_h(x,y,t)]$ with $h$ variables, where $x \in \R^n$, $y \in \R^m$, $t \in \R^{T}$. Alternatively, $T=s+1$ and $u = [\mathbb{X}]_{i-s}^i \in \R^{n \times m \times h \times (s+1)}$. 
        \STATE \textbf{Require} Candidate terms $\Phi(u)=\{\phi_1(u), \ldots, \phi_p(u)\}$ and corresponding coefficients $\Xi=\{\xi_1, \ldots, \xi_p\}$; the latent force model $Q_\pi (x,y,t)$ parameterized by $\pi$; the dynamics model $f_\theta(x,y,t)$ parameterized by $\theta$; pretrained weather forecasting model $g_\omega$ parameterized by $\omega$.
        \WHILE{not convergence}
            \STATE Optimize $\theta$, $\Xi$ and $\pi$ following Eqs. (\ref{eq:def1}-\ref{eq:def5});
        \ENDWHILE
        
        \STATE Generate super-resolution weather data by downscaling using $\mathbb{Y}=f_{\theta}(x',y',t)$, where $x' \in \R^{n'}, y' \in \R^{m'}$ are the coordinates of the super-resolution data with $n' > n, m' > m$, which are interpolations within the boundary of where the coordinates of $u$ are located;

        \WHILE{not convergence}
            \STATE Replace $[\mathbb{X}]_{i-s}^i$ by the augmented downscale data $\mathbb{Y}$ and optimize $\omega$ following Eqs. (\ref{eq:def6}-\ref{eq:def8});
        \ENDWHILE
        
        \STATE \textbf{Return} The well-trained weather forecasting model $g_\omega$.
    \end{algorithmic}
\end{algorithm}

For weather downscaling, the Huadong dataset consists of HRES and ERAs datasets. HRES represents a 10-day atmospheric model forecast, while ERA5 serves as a global atmospheric reanalysis, incorporating climate and weather observations. For regional downscaling, we construct a real-world dataset called "Huadong", covering the east China land and sea areas. In this dataset, HRES data is employed as the predictive data, while ERA5 reanalysis data serves as the ground truth. 

\textbf{Huadong dataset}: The Huadong dataset encompasses a latitude
range from $26.8^\circ N$ to $42.9^\circ N$ and a longitude range from $112.6^\circ E$ to $123.7^\circ E$. It comprises a grid of $64 \times 44$ cells, with each cell having a grid size of 0.25 degrees in both latitude and longitude. Notably, the Huadong dataset also incorporates Digital Elevation Model (DEM) data to represent terrain information. Since the terrain information usually refers to the boundary layers in the meteorology model instead of an individual weather factor in the PDE, for simplicity, we do not use this information in the paper. The HRES and ERA5 data cover the period from January 3, 2020, to April 1, 2022. The scores of the average of variables reported in Table \ref{tab:huadong} are computed based on all eight factors. Due to space limits, we only report the specific scores for four factors.

For weather forecasting, both Ningbo and Ningxia datasets consist of two main components: geographic data and meteorological data. The geographic data includes latitude, longitude, and DEM (Digital Elevation Model) information. The DEM information is commonly used in geographic information systems to represent the terrain of the area. On the other hand, the meteorological data in these datasets consist of various weather factors. These factors typically include wind speed, temperature, and pressure. These data provide information about the atmospheric conditions at different locations within the study area. To organize and represent the data, a grid format is used. In this format, the study area is divided into grids, and each grid cell represents a specific location. Within each grid cell, both the geographic and meteorological data for that location are stored.

\textbf{Ningbo dataset}: The Ningbo dataset represents a coastal area spanning from latitude $28.85^\circ$N to $30.56^\circ$N and longitude $120.91^\circ$E to $122.29^\circ$E. It is divided into a grid system comprising 58 grids in the latitude direction and 47 in the longitude direction. Each grid has a size of 0.03 degrees in both latitude and longitude. The DEM data are collected from ETOPO1\footnote{\href{https://www.ncei.noaa.gov/products/etopo-global-relief-model}{ETOPO1}}. The meteorological data are collected from Ningbo Meteorological Bureau\footnote{\href{http://zj.cma.gov.cn/dsqx/nbsqxj/}{Ningbo Meteorological Bureau}}, including 10 weather factors from 1/Jan/2021 to 1/Apr/2021 with 1-hour sample rate. Therefore, this dataset is real measurement data, rather than a typical reanalysis dataset.

\textbf{Ningxia dataset}: The Ningxia dataset represents a mountainous area spanning from latitude $34.5^\circ$N to $42^\circ$N and longitude $106^\circ$E to $116^\circ$E. There are 30$\times$40 grids with a grid size of 0.25 degrees in both latitude and longitude. The DEM data are collected from ETOPO1. The meteorological data are collected from ECMWF’s ERA5\footnote{\href{https://cds.climate.copernicus.eu/}{ECMWF’s ERA5}}, including 8 weather factors from 1/Jan/2021 to 1/Dec/2021 with 1-hour sample rate. For variable temperature, the atmosphere level is 500.   

\textbf{WeatherBench dataset}: The WeatherBench dataset\footnote{\url{https://mediatum.ub.tum.de/1524895}} represents a global weather dataset. We select the version with 5.625 degrees resolution, sampling rate of 6 hours, from 2008 to 2018, with five weather factors. This dataset originates from ECMWF’s ERA5 as well. For variable $z$ (geopotential), the atmosphere level is 500. For variable $t$ (temperature), the atmosphere level is 850.   

In the forecasting experiments, we divide each dataset into train, validation, and test sets using an
8:1:1 ratio in chronological order. The scores of the average of variables reported in Table \ref{tab:7dayningxia} are computed based on all eight factors. Due to space limits, we only report the specific scores for four factors.

\subsection{Experimental details}
We choose the learning rate at 1e-4, batch size of \textit{PhyDL-NWP} at 10000, and hidden dimension at 100. We perform grid search in [1e-5, 1e-4, 1e-3, 1e-2, 0.1, 1, 10] for the following hyperparameters and select them as: $\alpha$ at 10, $\beta$ at 1e-2. 

The neural network used for $f_\theta$ and $Q_\pi$ both consists of 8 hidden layers, where each layer has 100 neurons. MLP is the most commonly used model architecture in the literature of PINN, which leverages the universal function approximation capabilities to model complex mappings from inputs like (x, y, t) to outputs such as weather variables. This allows inference at any arbitrary resolution because the model outputs values at any real-valued coordinate, not restricted to the training grid.

We employ both automatic differentiation and finite difference (FD) methods in this paper. Specifically, once the neural network $f_\theta$ accurately approximates the weather data, we leverage PyTorch’s automatic differentiation to compute derivatives. PyTorch constructs a dynamic computational graph during forward operations, where each tensor records its computation history. When computing the output $y = f_\theta(t)$, invoking y.backward() enables access to the derivative with respect to $t$ via t.grad.

The finite difference (FD) method, on the other hand, is much simpler, but also much quicker. It approximates derivatives using discrete points. For the first derivative of a function \( u(t) \), there are:

\subsubsection{Forward Difference}
\[
\frac{du}{dt} \approx \frac{u(t+\Delta t) - u(t)}{\Delta t} + \mathcal{O}(\Delta t)
\]

\subsubsection{Backward Difference}
\[
\frac{du}{dt} \approx \frac{u(t) - u(t-\Delta t)}{\Delta t} + \mathcal{O}(\Delta t)
\]

\subsubsection{Central Difference}
\[
\frac{du}{dt} \approx \frac{u(t+\Delta t) - u(t-\Delta t)}{2\Delta t} + \mathcal{O}(\Delta t^2)
\]

We adopt the finite difference (FD) method to compute derivatives in the weather forecasting model $g_\omega$ (Eq.~\ref{eq:def8}), as automatic differentiation is not applicable—$g_\omega$ does not explicitly take time $t$ as an input, unlike $f_\theta(x, y, t)$. Another advantage is that first-order finite difference (FD) is much faster than automatic differentiation (auto-diff) because FD only requires basic arithmetic on neighboring grid points.
Auto-diff has to traverse the computation graph and involves memory-heavy backward passes, especially for deep networks or large spatiotemporal inputs.

\subsubsection{Forecasting baselines}
\label{app-baseline}
\begin{itemize}
    \item PINN~\citep{raissi2019physics}: a physics-informed feedforward neural network that incorporates known physical laws (e.g., PDEs) as soft constraints during training to improve generalization and physical consistency.
    \item Bi-LSTM-T \cite{yang2022correcting}: a deep learning (DL) model that uses Bi-LSTM for weather prediction.
    \item Hybrid-CBA \cite{han2022short}: a hybrid DL model that combines CNN, LSTM, and attention models for weather forecasting and correction.
    \item ConvLSTM \cite{shi2015convolutional}: a hybrid DL model that extends LSTM with convolutional gates.
    \item PINO~\citep{li2024physics}: a physics-informed neural operator that learns solution operators for PDEs directly, enabling efficient inference and better scalability than traditional PINNs.
    \item AFNO \cite{guibas2021adaptive, pathak2022fourcastnet}: a DL model that adapts Fourier neural operator for spatio-temporal modeling.
    \item MTGNN \cite{wu2020connecting}: a DL model that learns multivariate time series with graph neural networks.
    \item MegaCRN \cite{jiang2023spatio}: a deep learning model that learns heterogeneous spatial relationships with adaptive graphs. 
    \item ClimaX~\citep{nguyen2023climax}: a transformer-based deep learning foundation model for weather forecasting.
    \item FourcastNet~\citep{pathak2022fourcastnet}: a foundation model for global weather forecasting based on Adaptive Fourier Neural Operators.
    \item GraphCast~\citep{lam2022graphcast}: a foundation model for global weather prediction based on encoder-decoder with message passing.
\end{itemize}
Since most forecasting baselines are designed for single-step future prediction by default, we modify their neural architecture by multiplying the output dimension of the second-last layer (usually at the end of an LSTM or Conv block, before passing through the feed-forward network at the end) by the number of prediction steps. Since different baselines may operate on different resolutions of data, we will use interpolation to map to the desired resolution.

\subsubsection{Downscaling baselines}
\begin{itemize}
    \item Bicubic interpolation: a two-dimensional interpolation technique that uses the values and gradients of the function at surrounding grid points to obtain a smooth and continuous interpolated result. 
    \item FSRCNN \cite{passarella2022reconstructing}: a widely recognized method in computer vision, leveraged for both downscaling and single-image super-resolution, which conducts feature mapping using multi-layer CNNs and executes upsampling via deconvolution layers.
    \item ResDeepD \cite{sharma2022resdeepd}: a deep model that begins with an upsampling of the input to increase dimensions before proceeding to feature mapping via ResNet.
    \item EDSR \cite{jiang2022efficient}: a deep model that first conducts feature mapping using ResNet and then performs upsampling.
    \item RCAN \cite{yu2021deep}: a deep model based on ResNet that incorporates a global pooling layer for channel attention.
    \item YNet \cite{liu2020climate}: a novel deep convolutional neural network (CNN) with skip connections and fusion capabilities to perform downscaling for climate variables.
    \item DeepSD \cite{vandal2017deepsd}: a generalized stacked super resolution convolutional neural network (SRCNN) framework for statistical downscaling of climate variables.
    \item GINE~\citep{park2022downscaling}: a computer vision-based technique using topography-driven spatial and local-level information for downscaling climate simulation.
\end{itemize}